\documentclass[11pt,a4paper]{article}
\usepackage[hyperref]{emnlp2020}
\usepackage{times}
\usepackage{latexsym}
\usepackage{amsmath}
\usepackage{algorithm}
\usepackage{graphicx}
\usepackage{subcaption}
\usepackage[noend]{algpseudocode}
\usepackage{booktabs}
\usepackage{xcolor}
\usepackage{multirow}

\usepackage{microtype}

\aclfinalcopy %
\def\aclpaperid{*} %

\title{Incremental Neural Coreference Resolution in Constant Memory}

\author{Patrick Xia\textsuperscript{1}~~~~
  Jo\~ao Sedoc\textsuperscript{2}~~~~ Benjamin Van Durme\textsuperscript{1} \\
  \texttt{paxia@cs.jhu.edu}~~~\texttt{jsedoc@nyu.edu}~~~ \texttt{vandurme@cs.jhu.edu}\\
  \textsuperscript{1}Johns Hopkins University~~~~\textsuperscript{2}New York University}

\date{}

\begin{document}
\maketitle
\begin{abstract}
We investigate modeling coreference resolution under a fixed memory constraint by extending an incremental clustering algorithm to utilize contextualized encoders and neural components. Given a new sentence, our end-to-end algorithm proposes and scores each mention span against explicit entity representations created from the earlier document context (if any). These spans are then used to update the entity's representations before being forgotten; we only retain a fixed set of salient \textit{entities} throughout the document. In this work, we successfully convert a high-performing model \cite{joshi2019spanbert}, asymptotically reducing its memory usage to constant space with only a 0.3\% relative loss in F1 on OntoNotes 5.0. 

\end{abstract}

\section{Introduction}
\label{sec:introduction}

Coreference resolution is a core task in NLP for both model analysis and information extraction. At the sentence level, ambiguities in pronoun coreference can be used to probe a model for common sense \cite{levasque2012winograd, sakaguchi2019winogrande} or gender biases \cite{rudinger-etal-2018-gender, zhao-etal-2018-gender}. At the document level, coreference resolution is commonly used in information extraction pipelines, but can be applied to reading comprehension \cite{dasigi-etal-2019-quoref} or  literature analysis \cite{bamman-etal-2014-bayesian}.

Models for this task typically encode the entire text before scoring and subsequently clustering candidate mention spans, either found by a parser \cite{clark-manning-2016-improving} or learned jointly \cite{lee-etal-2017-end}. Prior work has primarily focused on improving pairwise span scoring functions \cite{raghunathan-etal-2010-multi, clark-manning-2016-deep, wu-etal-2020-corefqa} and methods for decoding into globally consistent clusters \cite{wiseman-etal-2016-learning, lee-etal-2018-higher, kantor-globerson-2019-coreference,  xu2020revealing}. Recent models have also benefited from pretrained encoders used to create high-dimensional input text (and span) representations, and improvements in contextualized encoders appear to translate directly to coreference resolution \cite{lee-etal-2018-higher, joshi-etal-2019-bert, joshi2019spanbert}. 

These models typically rely on simultaneous access to all spans -- $\Theta(n)$ for a document with length $n$ -- for \emph{scoring} and all scores -- up to $\Theta(n^2)$ -- for \emph{decoding}. As the dimensionality of contextualized encoders, and therefore the size of span representations, increases, this becomes computationally intractable for long documents or under limited memory. Given these constraints, expensive scoring functions are increasingly difficult to explore. Further, prior models depart from how humans incrementally read and reason about coreferent mentions; 
\citet{webster-curran-2014-limited} argue in favor of a limited memory constraint as a more psycholinguistically plausible approach to reading and model coreference resolution via shift-reduce parsing.

Motivated by scalability and armed with advances in neural architectures, we revisit that intuition. Following prior work, our model begins with a SpanBERT encoding of a text segment to form a list of proposed mention spans \cite{joshi-etal-2019-bert,joshi2019spanbert}. Clustering is performed online: each span either attaches to an existing cluster or begins a new one. We substantially minimize memory usage during inference by storing only the embeddings of active entities in the document and a small set of candidate mention spans. Our two contributions of online clustering and storing a constant size set of active entities result in an end-to-end trainable model that uses $O(1)$ space with respect to document length while sacrificing little in performance (see \autoref{fig:memory}).\footnote{Code and models available at \url{https://nlp.jhu.edu/incremental-coref}.}

\section{Model}
\label{sec:model}

Our algorithm revisits the approach taken by \citet{webster-curran-2014-limited} for incrementally making coreference resolution decisions (online clustering). The major differences lie in explicit entity representations, neural components, and learning. 

\paragraph{Baseline}
First, we summarize the coreference resolution model described by \citet{joshi-etal-2019-bert}, which itself extends from earlier work \cite{lee-etal-2017-end, lee-etal-2018-higher}. For each document, this model enumerates and scores all spans up to a chosen width. The span representations are formed using BERT \cite{devlin-etal-2019-bert} encodings of input text by concatenating the first, last, and an attention-weighted average of the token representations within the span. These spans are ranked and pruned to the top $\Theta(n)$ mentions. Both the maximum span width and fraction of remaining spans are hyperparameters. For each remaining span, the model learns a distribution over its possible antecedents (via a pairwise scorer) and the training objective maximizes the probability of its gold labeled antecedents. The entire model (including finetuning the encoder) is trained end-to-end over OntoNotes 5.0. 

This model is further improved by \citet{joshi2019spanbert}, who introduces SpanBERT and uses it as the underlying encoder instead. The SpanBERT-large version of  \citet{joshi-etal-2019-bert} is the baseline model used in this paper.

\paragraph{Inference}
Our method (Algorithm \ref{alg:inference}) stores a permanent list of entities (clusters), each with its own representation. For a given sentence or segment, the model proposes a candidate set of spans. For each span, a \textit{scorer} scores the \textit{span} representation against all the \textit{cluster} representations. This is used to determine to which (if any) of the pre-existing clusters the current span should be added. Upon inclusion of the span in the cluster, the cluster's representation is subsequently updated via a (learned) function. Periodically, the model evicts less salient entities, writing them to disk. Under this algorithm, each clustering decision is permanent.\footnote{This uses greedy decoding; exploring decoding strategies is beyond the scope of this work, which is focused on memory.} 

\begin{algorithm}
\small
\caption{FindClusters(Document)}
\label{alg:inference}
\begin{algorithmic}
\State Create an empty Entity List, $E$
\For{segment $\in$ Document}
    \State $M \gets$ \textsc{Spans}$($segment$)$
    \For {$m \in M$}
        \State $scores \gets \textsc{PairScore}(m, E)$
        \State $top\_score \gets \text{max}(scores)$ 
        \State $top\_e \gets \text{argmax}(scores)$
        \If {$top\_score > 0$}
            \State $\textsc{Update}(top\_e, m)$
        \Else
            \State $\textsc{Add\_new\_entity}(E, m)$
        \EndIf
    \EndFor
    \State \textsc{Evict}($E$)
\EndFor
\State \Return $E$
\end{algorithmic}
\end{algorithm}

Concretely, our model uses a contextualized encoder, SpanBERT \cite{joshi2019spanbert}, to encode an entire segment. Given a segment, \textsc{Spans} returns candidate spans, a result of enumerating all spans up to a fixed width, encoding spans as a combination of the embeddings within the span, and pruning using a learned scorer, following prior work \cite{lee-etal-2017-end, joshi-etal-2019-bert}.

\textsc{PairScore} is a feedforward scorer which takes as input the concatenation of a mention span and entity representation along with additional embeddings for distance and genre. \textsc{Update} updates the entity representation ($\mathbf{e}_{top\_e}$) with the newly linked span representation ($\mathbf{e}_{m}$). In this work, we use a learned weight, $\alpha = \sigma(\text{FF}([\mathbf{e}_{top\_e}, \mathbf{e}_{m}]))$ and update $\mathbf{e}_{top\_e} \leftarrow \alpha\mathbf{e}_{top\_e} + (1-\alpha)\mathbf{e}_{m}$.\footnote{Using a simple moving average performs slightly worse.} Here, FF is a feedforward network and $\sigma$ is the sigmoid function.

To ensure constant space, \textsc{Evict} moves some entities from $E$ to CPU. These entities are never revisited; the offsets are stored on CPU solely for evaluation purposes. We evict based on cluster size and distance from the end of the segment.

The algorithm is independent of these components, so long as they satisfy the correct interface. Specifically, our algorithm is compatible with the recent model by \citet{wu-etal-2020-corefqa}. They use a query-based pairwise scorer, which could be adopted in place of the feedforward pairwise scorer. Our use of abstract components also allows for comparison of different encoders or update rules.

\begin{table*}[ht!]
    \small
    \centering
    \begin{tabular}{lcccccccccc}
    \toprule
     & \multicolumn{3}{c}{MUC} & \multicolumn{3}{c}{B$^3$} & \multicolumn{3}{c}{CEAF$_{\phi_4}$} & \\
    & P & R & F1 & P & R & F1 & P & R & F1 & Avg. F1\\
    \midrule
     Baseline \cite{joshi2019spanbert} & 85.8 & 84.8 & 85.3 & 78.3 & 77.9 & 78.1 & 76.4 & 74.2 & 75.3 & 79.6\\
     Ours & 85.7 & 84.8 & 85.3 & 78.1 & 77.5 & 77.8 & 76.3 & 74.1 & 75.2 & 79.4\\
     Ours (without eviction) & 85.7 & 84.9 & 85.3 & 78.1 & 77.5 & 77.8 & 76.2 & 74.2 & 75.2 & 79.4\\
     \midrule
    CorefQA \cite{wu-etal-2020-corefqa} & 88.6 & 87.4 & 88.0 & 82.4 & 82.0 & 82.2 & 79.9 & 78.3 & 79.1 & 83.1 \\
    \bottomrule
    \end{tabular}
    \caption{Complete results of our model on the OntoNotes 5.0 test set with three coreference resolution metrics: MUC, B$^3$, and CEAF$_{\phi_4}$. For completeness, we also present the values for the current state-of-the-art. All models use an encoder derived from SpanBERT-large.}
    \label{tab:full}
\end{table*}

\paragraph{Training}
Similar to prior work \cite{lee-etal-2017-end}, our training objective is to maximize the probability of the correct antecedent (cluster) for each mention span. However, rather than considering \textit{all} correct antecedents, we are only interested in the cluster for the \textit{most recent} one.\footnote{Scoring is between mention spans and entity clusters, so there needs to be a single correct cluster.} For each mention $m$, $scores$ is treated as an unnormalized probability distribution $P(e \mid m)$ for $e \in E$, where $E$ is the entity list that includes an $\varepsilon$ target label which represents the action of starting a new cluster. The exact objective is to maximize $P(e = e_{\text{gold}} \mid m)$; $e_{\text{gold}}$ is the gold cluster of $m$ (i.e., the cluster the most recent antecedent was assigned to).

However, the entirely sequential algorithm also introduces sample inefficiency, as most mentions have the same label ($\varepsilon$) and barely accrue loss. We speed up training by accumulating gradients periodically, trading computation time for space. 
This tradeoff is similar to that of batching by documents, which is impractical for our model from a memory perspective. Like prior work, we update parameters once per document (and not once per mention).

We lean on pretrained components: we reuse not only encoder weights that are already finetuned on this dataset, but also the mention and pairwise scorers from \citet{joshi2019spanbert} as initialization for our encoder, \textsc{Spans} and \textsc{PairScore}.\footnote{The implementation of \newcite{joshi2019spanbert,joshi-etal-2019-bert} was the most amenable to extension and experimentation and therefore serves as our illustrative example.} %

\section{Experiments}
\label{sec:exps}

Since we reuse weights from \citet{joshi2019spanbert} (our baseline), our primary experiment is to compare their model to our constant space adaptation in both task performance and memory usage. Additionally, we analyze document and segment length, conversational genre, and explicit clusters.

\paragraph{Data} We use OntoNotes 5.0 \cite{weischedel2013ontonotes, pradhan2013towards}, which consists of 2,802, 343, and 348 documents in the training, development and test splits respectively. These documents span several genres, including those with multiple speakers (broadcast and telephone conversations) and those without (broadcast news, newswire, magazines, weblogs, and the Bible). 

\paragraph{Implementation} We use the model dimensions and training hyperparameters from the baseline model, a publicly available coreference resolution model by \citet{joshi-etal-2019-bert, joshi2019spanbert}. We also reuse their (trained) parameters for the encoder, span scorer, and span pair scorer as initialization. However, our model does not make use of speaker features, since it is not meaningful to assign a speaker to the cluster representation. At the end of each segment, we evict singleton (size 1) clusters more than 600 tokens away from the end of the segment. Additionally, we evict all clusters whose most recent member is more than 1200 tokens away. In this work, we also freeze the encoder---further finetuning the encoder provided little, if any, benefit likely because the encoder has already been finetuned on this dataset and task. Additional details, including our choice of eviction function, are described in \autoref{sec:appendix:hparams}. All experiments are performed on either a single NVIDIA 1080 TI (11GB) or GTX Titan X (12GB).

\section{Results}
\label{sec:results}

\subsection{Performance}
\label{sec:results:perf}

\autoref{tab:full} presents the OntoNotes 5.0 test set scores for the metrics: MUC \cite{vilain-etal-1995-model}, B$^3$ \cite{Bagga98algorithmsfor}, and CEAF$_{\phi_4}$ \cite{luo-2005-coreference} using the official CoNLL-2012 scorer. We reevaluated the baseline, and we report the scores for CorefQA directly from \citet{wu-etal-2020-corefqa}. We observe a small drop in performance compared to the baseline and apparently no drop with eviction.

\begin{table}[ht]
    \small
    \centering
    \begin{tabular}{rccccr}
    \toprule
     Subset & \#Docs & \textbf{JS-L} & Ours & $\Delta$ & -evict\\
    \midrule
     All & 343 & 80.1 & 79.5 & -0.6 & 79.7\\
     0-128 & 57 & 84.6 & 84.5 & -0.1 & 84.5 \\
     129-256 & 73  & 83.7 & 83.6 & -0.1 & 83.6 \\
     257-512 & 78  & 82.9 & 83.4 & +0.5 & 83.4\\
     513-768 & 71  & 80.1 & 79.3 & -0.8 & 79.3 \\
     769-1152 & 52  & 79.1 & 78.6 & -0.5 & 79.0 \\
     1153+ & 12  & 71.3 & 69.6 & -1.7 & 69.8\\
     \midrule
     1 Speaker & 268 & 81.1 & 81.0 & -0.1 & 81.2\\
     2+ Speakers & 75 & 76.7 & 75.0 & -1.7 & 75.0\\
     \midrule
     Test & 348 & 79.6 & 79.4 & -0.2 & 79.4\\
    \bottomrule
    \end{tabular}
    \caption{Average F1 score on the development set broken down by document length and number of speakers. \textbf{JS-L} refers to the \texttt{spanbert\_large} model from \citet{joshi2019spanbert}, which we treat as our baseline, and -evict refers to the model without eviction.}
    \label{tab:distance}
\end{table}

\subsection{Document Length}

Our goal is a constant-memory model that is comparable to the baseline. We showed above that our model is competitive with and without eviction, the key to constant memory. In \autoref{tab:distance}, we report the average F1 broken down based on the length (in subtokens)\footnote{This split of the development set differs from that used by \citet{joshi-etal-2019-bert} which counts the number of 128-subtoken sized segments. We directly count subtokens.} of the document and number of speakers. Our model is competitive on most document sizes and in the single speaker setting. On longer documents, eviction has a minor effect. Because our model does not make use of speaker embeddings, we perform worse on documents with multiple speakers. This drop due to speaker features matches previous findings \cite{lee-etal-2017-end}. One way to include speakers and retain speaker-independent entity embeddings is by treating speakers as part of the input text \cite{wu-etal-2020-corefqa}.

\subsection{Inference Memory}
\label{sec:results:memory}

\begin{table}[t]
    \small
    \centering
    \begin{tabular}{rcc}
    \toprule
     Model & GPU Memory (GB) & Dev. F1 \\
    \midrule
     Our model & 2.0 & 79.5 \\
     No eviction & 2.0 & 79.7 \\
     \midrule 
     \textbf{JS-B} & 6.4 & 77.7 \\
     \textbf{JS-L} & \textgreater 11.9 & 80.1 \\
     \bottomrule
    \end{tabular}
    \caption{Space needed and performance over the development set. \textbf{JS-B} and \textbf{JS-L} refer to the \texttt{base} and \texttt{large} variants SpanBERT used in the baseline.}
    \label{tab:memory}
\end{table}

We now look towards space. In \autoref{tab:memory}, we report the space needed to perform inference over the entire development set. Compared to the baseline and its smaller \texttt{base} version, our model uses substantially less memory. We also find that eviction has little effect on memory and F1 on this dataset. 

\begin{figure}
    \centering
    \def\svgwidth{\columnwidth}
    \input{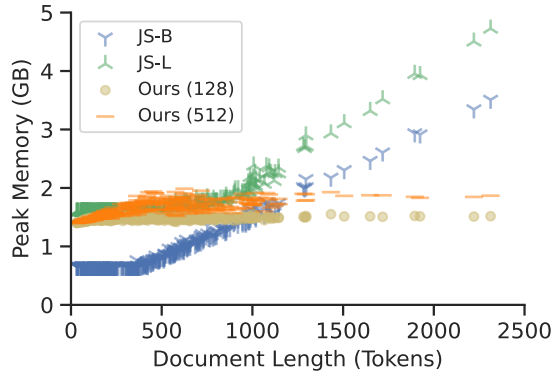}
    \caption{Total size of GPU-allocated tensors for each document in the development set. The base (\textbf{JS-B}) and large (\textbf{JS-L}) models of the baseline use apparently linear space, while ours with inference segment lengths of 128 and 512 use constant space.}
    \label{fig:memory}
\end{figure}

Usage in practice is subject to the memory allocator, and our implementation (PyTorch) differs in framework from the baseline (TensorFlow). To fairly compare the two models, we compute the maximum \textit{space used by the allocated tensors} for each document during inference.\footnote{For profiling, we use \texttt{run\_op\_benchmark} for TensorFlow 1.15 and \texttt{pytorch\_memlab} 0.0.4 and \texttt{torch.cuda} for PyTorch 1.5.} \autoref{fig:memory} compares this value of peak theoretical memory usage of several models against the dataset. It shows the baseline is dominated by a term that grows linearly with length, while that is not the case for our model, which has constant space usage.

Our model reduces the asymptotic memory usage to $O(1)$. In addition, these plots do not clearly show asymptotic memory usage: the baseline and other derivative models have a quadratic component for scoring span pairs (with a small coefficient). The encoder, SpanBERT, adds a significant constant term (with respect to document length) to all models. While there is some work in sparsifying Transformers \cite{child2019sparsetransformer, Kitaev2020Reformer:}, there does not yet exist a sparse SpanBERT.

These plots show that models have relatively modest memory usage during inference. However, their usage grows in training, due to gradients and optimizer parameters. This additional memory usage would render training and finetuning the underlying encoder infeasible for the baseline but possible using our model with 12GB GPUs.

\begin{figure*}[ht]
    \centering
    \def\svgwidth{0.97\columnwidth}
    \begin{subfigure}{0.97\columnwidth}
    \vspace{2mm}
        \input{figures/clusters_tex.pdf_tex}
    \end{subfigure}
    \begin{subfigure}{\columnwidth}
    \fbox{
    \parbox{0.93\columnwidth\linespread{1.0}\selectfont}{
        \begin{small}
        \textit{
            President Clinton may travel to North Korea in an attempt to improve relations with that country. The announcement comes after two days of talks between American and North Korean leaders in Washington. Secretary of State Madeleine Albright has accepted an invitation to visit North Korea and meet with leader Kim Jong-il. She made the unexpected announcement at a dinner last night in Washington. North Korea's top defense official hosted the event. The country is on a U.S. list of nations that sponsor terrorism. The Clinton administration is trying to persuade North Korea to halt its ballistic missile program as a way it can get off the list. There's no word yet when Albright's trip will take place.
        }
        \end{small}
        }
    }
    \end{subfigure}
    \caption{t-SNE plot (left) of span representations of a single document (right) in the development set (cnn\_0040\_0). Each color/shape is a predicted cluster, while light gray circles indicate predicted singletons. For each span, the gold cluster label (-1, if not annotated) and its contribution to the entity embedding is noted in parentheses.}
    \label{fig:clusters}
\end{figure*}

\subsection{Segment Length}
\label{sec:results:seglen}

The memory usage at each step (and therefore of the algorithm) is also dependent on the segment length due to the encoder. \autoref{tab:segment} explores the effect of the length of each segment (split at sentence boundaries), which gives us further insight into the tradeoff between performance and memory reduction. We compare models without eviction to ensure fairness. Our observations follow those from \citet{joshi-etal-2019-bert} that larger context windows compatible with the encoder input size improve performance. We also observe that models trained on shorter sequences can be scaled, at inference time, to longer sequences and obtain gains in performance. There is an unsurprising substantial drop using single sentences, owing to coreference being a cross-sentence phenomenon.

\addtolength{\tabcolsep}{-3pt}
\begin{table}[ht]
    \small
    \centering
    \begin{tabular}{crcccc}
    \toprule
    & & \multicolumn{4}{c}{Inference Length}\\
     & & \multicolumn{2}{c}{Sentences} & \multicolumn{2}{c}{Tokens} \\
     &Train$\downarrow$ & 1 sent. & 10 sent. & 128 toks. & 512 toks. \\
      \midrule
     \multirow{3}{*}{\rotatebox[origin=c]{90}{sents.}} & 1 & 70.0 & 76.4 & 75.2 & 76.9 \\
     & 5 & 70.0 & 77.4 & 76.4 & 78.6 \\
     & 10 & 68.9 & 77.8 & 76.2 & 78.9\\
     \midrule
     \parbox[t]{2mm}{\multirow{4}{*}{\rotatebox[origin=c]{90}{toks.}}} & 128  & 70.1 & 77.2 & 76.3 & 77.7 \\
     & 256  & 69.1 & 77.9 & 76.5 & 78.8 \\
     & 384  & 67.7 & 77.3 & 76.1 & 79.1 \\
     & 512  & 67.1 & 77.7 & 75.6 & 79.7 \\
     \bottomrule
    \end{tabular}
    \caption{Average dev. F1 score for models trained and evaluated across a range of segment lengths (either fixed number of sentences or subtokens).}
    \label{tab:segment}
\end{table}
\addtolength{\tabcolsep}{3pt}

\subsection{Span Representations}
\label{sec:results:spans}

\autoref{fig:clusters} visualizes the proposed span representations for a single document in the development set. The colors/shapes represent our predictions, and each point is annotated with the text, the gold cluster label, and the (normalized) $\alpha$ for each span (recall $\alpha$ is used in the \textsc{Update} function to determine a span's contribution to its entity embedding). 

Given these embeddings, the figure supports the viability of clustering approaches: gold coreference clusters  tend to be ``close'' in embedding space. Regarding $\alpha$, some spans are weighted equally (``Clinton'') while others are not (``North Korea''). This could be a result of online updates biasing more recent spans with higher weights. Alternatively, it may suggest that some spans (like names) are more informative than others (like pronouns). %

\section{Conclusion}
\label{sec:conclusion}

We present an online algorithm for space efficient coreference resolution that incorporates contributions from recent neural end-to-end models. We show it is possible to transform a model which performs document-level inference into an incremental algorithm. In so doing, we greatly reduce the memory usage of the model during inference at virtually no cost to performance, thereby providing an option for researchers and practitioners interested in modern coreference resolution models for tasks constrained by memory, like the modeling of book-length texts.

\section*{Acknowledgments}        
We would like to thank Aaron White for helpful discussions. This work was supported in part by DARPA AIDA (FA8750-18-2-0015) and IARPA BETTER (\#2019-19051600005). The views and conclusions contained in this work are those of the authors and should not be interpreted as necessarily representing the official policies, either expressed or implied, or endorsements of DARPA, ODNI, IARPA, or the U.S. Government. The U.S. Government is authorized to reproduce and distribute reprints for governmental purposes notwithstanding any copyright annotation therein.

\bibliography{emnlp2020}
\bibliographystyle{acl_natbib}

\appendix
\section{Hyperparameters}
\label{sec:appendix:hparams}

In this section, we describe several implementation details and other experiments that we tried. To improve memory usage, we use gradient accumulation. Ultimately, all training was performed on the NVIDIA 1080 TI (11GB), on which we accumulate gradients when the memory usage exceeds 7.5GB. In initial trials, we explored sampling losses for negative examples (spans that do not have an antecedent). While we found sampling at a rate of 0.2 (for example) would speed up training and inference, ultimately it contributed up to a one point deficit in F1. 

We also explored teacher forcing, in which spans are added to the gold cluster during training instead of the predicted one. This would ``correct'' the training objective to match prior work. However, this did not have a noticeable effect on performance. Likewise, we were able to train a competitive model for which only the SpanBERT encoder from \citet{joshi-etal-2019-bert} was retained and the span scorer and pairwise scorer were randomly initialized. However, we opted not to use that for the full experiments because training was more expensive in time. Further, learning span detection is not guaranteed by this objective, leading to high variance across runs (most notably in the number of epochs). Thus, the effect of other hyperparameters would not be immediately apparent. 

Additionally, we attempted further finetuning the encoder with a separate learning rate of [1e-5, 5e-6], but were unsuccessful in improving the performance. On our GPUs, training (without finetuning) roughly takes 70 min/epoch with negative sample rate 0.2, 100 min/epoch without sampling loss, and 160 min/epoch when finetuning. All runs are stopped after 5 to 15 epochs due to early stopping (patience = 5). 

For eviction, a policy which evicts singletons distance $>600$ and all clusters distance $>1200$ would have a recall of 99.57\% over the training set. This is a result of sweeping over [200, 300, 400, 500, 600, 900] for singletons and [400, 600, 800, 1000, 1200, 1800] for all clusters. We also try using a single fixed distance, as well as other non-constant schemes (e.g. size $\times$ distance as thresholds). Here, distance is between the current point in the document and the average of the start and end indices of the most recent span added to the cluster. We selected this policy from several other choices due to the recall it achieved.

Our model dimensions otherwise match up exactly with \citet{joshi-etal-2019-bert}. Rather than omitting the speaker embedding and segment length embedding entirely (which would affect pairwise scorer dimensionality), we replace those embeddings with the zero vector.

Concretely, we performed grid searches over dropout ([0.3, 0.4, 0.5]), sample rate ([0.2, 0.5, 0.75, 1.0]), and update method ([alpha, mean]). We find that 0.4 dropout, 1.0 sample rate, and alpha weighting were the best after 2 epochs. Alpha weighting resulted in, on average, approximately 0.1 F1 improvement (after 2 epochs). 

For alpha weighting, we used a two-layer MLP: the first layer has size 300 and ReLu nonlinearity, while the final layer then projected to a scalar with a sigmoid activation. After fixing those values, we explored learning rate ([5e-5, 1e-4, 2e-4, 5e-4]), eviction policy at training ([no eviction, eviction]), and gradient clipping value ([1, 5, 10]). Here, we found that 2e-4, no eviction, and gradient clipping at 10 performed slightly better, although there was little difference between them after these models were allowed to converge. 

Given the final set of hyperparameters, we performed five training runs, resulting in average development set F1 of [79.4, 79.5, 79.5, 79.5, 79.7]. We selected the best performing model for the results in the paper. For \autoref{tab:segment}, we trained each model only once. 

For these experiments, our model contains 377M parameters, of which 340M is SpanBERT-large \cite{joshi2019spanbert}.

\end{document}